\documentclass[sigconf,edbt]{acmart-edbt2019}

\PassOptionsToPackage{hyphens}{url} % specify option hyphens for url package to not use hyphens in break.
\usepackage{hyperref} % automatically hyperlinks from the table of contents to the sections, to equations, etc. and also tweak PDF metadatas. This also loads package url.
\usepackage[section]{placeins}

\usepackage[hang]{footmisc}  % option hang: alignment
\usepackage{lipsum} % for generating dummy text.

\usepackage[disable]{todonotes} % disable all notes.

\usepackage[T1]{fontenc} % font format on the output pdf.
\usepackage[utf8]{inputenc} % encoding of content of .tex file.

\usepackage{graphicx} % enhanced support for graphics.
\usepackage{subfigure}
\usepackage{tikz} % drawing language, gallery with examples: http://www.texample.net/tikz/examples/.

\usepackage{booktabs} % enhancements of the standard LaTeX tables.
\usepackage{multirow} % tables with cells spanning multiple rows.
\usepackage{multicol} % tables with cells spanning multiple columns.
\usepackage{makecell} % make cell with more control text format.
\usepackage{array, tabularx, boldline} % some easier table.
\usepackage{mparhack} % workaround for a bug on LaTeX \marginpar.
\usepackage[justification=centering]{caption}
\usepackage{amsthm}
\usepackage{amsfonts}
\usepackage{bm} % bold math font.
\usepackage{mathtools} % enhancements of the standard AMS-LaTeX packages.
\usepackage{mathdots} % redefines \dots and its friends so that they change size when appropriate.
\usepackage{xfrac} % to typeset nice inline fractions, as in ½.
\usepackage{faktor} % same as before, but without shrinking denominator’s and numerator’s size; useful for quotients.
\usepackage{cancel} % to draw diagonal bars above terms in an equation.

\usepackage{enumitem}

\def\eqref#1{equation~\ref{#1}}
\def\1{\bm{1}}

\def\vc{{\bm{c}}}

\def\ve{{\bm{e}}}

\def\vh{{\bm{h}}}
\def\vi{{\bm{i}}}
\def\vj{{\bm{j}}}
\def\vk{{\bm{k}}}

\def\vr{{\bm{r}}}

\def\vt{{\bm{t}}}

\DeclareMathAlphabet{\mathsfit}{\encodingdefault}{\sfdefault}{m}{sl}
\SetMathAlphabet{\mathsfit}{bold}{\encodingdefault}{\sfdefault}{bx}{n}

\def\gD{{\mathcal{D}}}

\def\gL{{\mathcal{L}}}

\def\gS{{\mathcal{S}}}

\def\sC{{\mathbb{C}}}

\def\sH{{\mathbb{H}}}

\def\sR{{\mathbb{R}}}

\setcopyright{rightsretained}
\renewcommand\footnotetextcopyrightpermission[1]{\footnote{Published in the Proceedings of the 1st International Workshop on Data Science for Industry 4.0 (DSI4) at EDBT/ICDT 2019 Joint Conference. Full version. \\
Copyright \textcopyright 2019 held by the owner(s)/author(s). Distribution of this paper is permitted under the terms of the Creative Commons license CC-BY-NC 4.0.}}

\acmDOI{}

\acmISBN{XXX-X-XXXXX-XXX-X}

\acmConference[EDBT 2019]{22nd International Conference on Extending Database Technology (EDBT)}{March 26-29, 2019}{Lisbon, Portugal} 
\acmYear{2019}

\begin{document}
% reduce list vspace
\setlist[itemize]{noitemsep, topsep=0pt}
\setlist[enumerate]{noitemsep, topsep=0pt}

% reduce equation vspace
%\setlength\abovedisplayskip{0pt}
\setlength\belowdisplayskip{0pt}
\setlength\belowdisplayshortskip{0pt}

% reduce table, figure vspace
%\setlength{\floatsep}{0pt}
%\setlength{\textfloatsep}{0pt}
%\setlength{\intextsep}{0pt}

\title{Analyzing Knowledge Graph Embedding Methods from a Multi-Embedding Interaction Perspective}
\author{Hung Nghiep Tran}
\affiliation{%
	\institution{The Graduate University for Advanced Studies, \\SOKENDAI}
	\country{Tokyo, Japan}
}
\email{nghiepth@nii.ac.jp}
\author{Atsuhiro Takasu}
\affiliation{%
	\institution{National Institute of Informatics}
	\city{Tokyo} 
	\country{Japan}
}
\email{takasu@nii.ac.jp}

\begin{abstract}
Knowledge graph is a popular format for representing knowledge, with many applications to semantic search engines, question-answering systems, and recommender systems. Real-world knowledge graphs are usually incomplete, so knowledge graph embedding methods, such as Canonical decomposition/Parallel factorization (CP), DistMult, and ComplEx, have been proposed to address this issue. These methods represent entities and relations as embedding vectors in semantic space and predict the links between them. The embedding vectors themselves contain rich semantic information and can be used in other applications such as data analysis. However, mechanisms in these models and the embedding vectors themselves vary greatly, making it difficult to understand and compare them. Given this lack of understanding, we risk using them ineffectively or incorrectly, particularly for complicated models, such as CP, with two role-based embedding vectors, or the state-of-the-art ComplEx model, with complex-valued embedding vectors. In this paper, we propose a multi-embedding interaction mechanism as a new approach to uniting and generalizing these models. We derive them theoretically via this mechanism and provide empirical analyses and comparisons between them. We also propose a new multi-embedding model based on quaternion algebra and show that it achieves promising results using popular benchmarks. \\\\Source code is available on GitHub at \\\url{https://github.com/tranhungnghiep/AnalyzeKGE}.
\end{abstract}

\keywords{Knowledge Graph, Knowledge Graph Completion, Knowledge Graph Embedding, Multi-Embedding, Representation Learning.}

\maketitle              % typeset the header of the contribution

\section{Introduction}
Knowledge graphs provide a unified format for representing knowledge about relationships between entities. A knowledge graph is a collection of triples, with each triple $ (h, t, r) $ denoting the fact that relation $ r $ exists between head entity $ h $ and tail entity $ t $. Many large real-world knowledge graphs have been built, including WordNet \cite{millergeorgea._wordnetlexicaldatabase_1995} representing English lexical knowledge, and Freebase \cite{bollacker_freebasecollaborativelycreated_2008} and Wikidata \cite{vrandecic_wikidatafreecollaborative_2014} representing general knowledge. Moreover, knowledge graph can be used as a universal format for data from applied domains. For example, a knowledge graph for recommender systems would have triples such as \textit{(UserA, Item1, review)} and \textit{(UserB, Item2, like)}.

Knowledge graphs are the cornerstones of modern semantic web technology. They have been used by large companies such as Google to provide semantic meanings into many traditional applications, such as semantic search engines, semantic browsing, and question answering \cite{amitsinghal_officialgoogleblog_2012}. One important application of knowledge graphs is recommender systems, where they are used to unite multiple sources of data and incorporate external knowledge \cite{carrer-neto_socialknowledgebasedrecommender_2012} \cite{zhang_collaborativeknowledgebase_2016}. Recently, specific methods such as knowledge graph embedding have been used to predict user interactions and provide recommendations directly \cite{he_translationbasedrecommendation_2017}.

Real-world knowledge graphs are usually incomplete. For example, Freebase and Wikidata are very large but they do not contain all knowledge. This is especially true for the knowledge graphs used in recommender systems. During system operation, \textit{users} \textit{review} new \textit{items} or \textit{like} new \textit{items}, generating new triples for the knowledge graph, which is therefore inherently incomplete. Knowledge graph completion, or link prediction, is the task that aims to predict new triples.

This task can be undertaken by using knowledge graph embedding methods, which represent entities and relations as embedding vectors in semantic space, then model the interactions between these embedding vectors to compute matching scores that predict the validity of each triple. Knowledge graph embedding methods are not only used for knowledge graph completion, but the learned embedding vectors of entities and relations are also very useful. They contain rich semantic information similar to word embeddings \cite{mikolov_distributedrepresentationswords_2013} \cite{mikolov_efficientestimationword_2013} \cite{jeffreypennington_gloveglobalvectors_2014}, enabling them to be used in visualization or browsing for data analysis. They can also be used as extracted or pretrained feature vectors in other learning models for tasks such as classification, clustering, and ranking.

Among the many proposed knowledge graph embedding methods, the most efficient and effective involve trilinear-product-based models, such as Canonical decomposition/Parallel factorization (CP) \cite{hitchcock_expressiontensorpolyadic_1927} \cite{lacroix_canonicaltensordecomposition_2018}, DistMult \cite{yang_embeddingentitiesrelations_2015}, or the state-of-the-art ComplEx model \cite{trouillon_complexembeddingssimple_2016}. These models solve a tensor decomposition problem with the matching score of each triple modeled as the result of a trilinear product, i.e., a multilinear map with three variables corresponding to the embedding vectors $ \vh $, $ \vt $, and $ \vr $ of head entity $ h $, tail entity $ t $, and relation $ r $, respectively. The trilinear-product-based score function for the three embedding vectors is denoted as $ \langle \vh, \vt, \vr \rangle $ and will be defined mathematically in Section \ref{sect:relatedwork}.
 
However, the implementations of embedding vectors for the various models are very diverse. DistMult \cite{yang_embeddingentitiesrelations_2015} uses one real-valued embedding vector for each entity or relation. The original CP \cite{hitchcock_expressiontensorpolyadic_1927} uses one real-valued embedding vector for each relation, but two real-valued embedding vectors for each entity when it is as head and as tail, respectively. ComplEx \cite{trouillon_complexembeddingssimple_2016} uses one complex-valued embedding vector for each entity or relation. Moreover, a recent heuristic for CP \cite{lacroix_canonicaltensordecomposition_2018}, here denoted as CP$ _h $, was proposed to augment the training data, helping CP achieve results competitive with the state-of-the-art model ComplEx. This heuristic introduces an additional embedding vector for each relation, but the underlying mechanism is different from that in ComplEx. All of these complications make it difficult to understand and compare the various models, to know how to use them and extend them. If we were to use the embedding vectors for data analysis or as pretrained feature vectors, a good understanding would affect the way we would use the complex-valued embedding vectors from ComplEx or the different embedding vectors for head and tail roles from CP.

In this paper, we propose a multi-embedding interaction mechanism as a new approach to uniting and generalizing the above models. In the proposed mechanism, each entity $ e $ is represented by multiple embedding vectors $ \{\ve^{(1)}, \ve^{(2)}, \dots\} $ and each relation $ r $ is represented by multiple embedding vectors $ \{\vr^{(1)}, \vr^{(2)}, \dots\} $. In a triple $ (h, t, r) $, all embedding vectors of $ h $, $ t $, and $ r $ interact with each other by trilinear products to produce multiple interaction scores. These scores are then weighted summed by a weight vector $ \bm{\omega} $ to produce the final matching score for the triple. We show that the above models are special cases of this mechanism. Therefore, it unifies those models and lets us compare them directly. The mechanism also enables us to develop new models by extending to additional embedding vectors.

In this paper, our contributions include the following.
\begin{itemize}
	\item We introduce a multi-embedding interaction mechanism as a new approach to unifying and generalizing a class of state-of-the-art knowledge graph embedding models.
	\item We derive each of the above models theoretically via this mechanism. We then empirically analyze and compare these models with each other and with variants.
	\item We propose a new multi-embedding model by an extension to four-embedding vectors based on quaternion algebra, which is an extension of complex algebra. We show that this model achieves promising results.
\end{itemize}

\section{Related Work} \label{sect:relatedwork}
Knowledge graph embedding methods for link prediction are actively being researched \cite{wang_knowledgegraphembedding_2017}. Here, we only review the works that are directly related to this paper, namely models that use only triples, not external data such as text \cite{wang_knowledgegraphtext_2014} or graph structure such as relation paths \cite{lin_modelingrelationpaths_2015}. Models using only triples are relatively simple and they are also the current state of the art.

\subsection{General architecture}
Knowledge graph embedding models take a triple of the form $ (h, t, r) $ as input and output the validity of that triple. A general model can be viewed as a three-component architecture: 
\begin{enumerate}
	\item \textit{Embedding lookup:} linear mapping from one-hot vectors to embedding vectors. A one-hot vector is a sparse discrete vector representing a discrete input, e.g., the first entity could be represented as $ [1, 0, \dots, 0]^\top $. A triple could be represented as a tuple of three one-hot vectors representing $ h $, $ t $, and $ r $, respectively. An embedding vector is a dense continuous vector of much lower dimensionality than a one-hot vector thus lead to efficient distributed representations \cite{hinton_learningdistributedrepresentations_1986} \cite{hinton_distributedrepresentations_1984}.
	\item \textit{Interaction mechanism:} modeling the interaction between embedding vectors to compute the matching score of a triple. This is the main component of a model.
	\item \textit{Prediction:} using the matching score to predict the validity of each triple. A higher score means that the triple is more likely to be valid.
\end{enumerate}

\subsection{Categorization}
Based on the modeling of the second component, a knowledge graph embedding model falls into one of three categories, namely \textit{translation-based}, \textit{neural-network-based}, or \textit{trilinear-product-based}, as described below.

\subsubsection{Translation-based:} These models translate the head entity embedding by summing with the relation embedding vector, then measuring the distance between the translated images of head entity and the tail entity embedding, usually by $ L^1 $ or $ L^2 $ distance:
\begin{equation} \label{eq:trans}
\begin{split}
\gS(h,t,r) = &-|| \vh + \vr - \vt ||_p \\
= &-\left(\sum_d^D |h_d + r_d - t_d|^p\right)^{1/p},
\end{split}
\end{equation}
where 
\begin{itemize}
	\item $ \vh, \vt, \vr $ are embedding vectors of $ h $, $ t $, and $ r $, respectively,
	\item $ p $ is 1 or 2 for $ L^1 $ or $ L^2 $ distance, respectively,
	\item $ D $ is the embedding size and $ d $ is the dimension for which $ h_d $, $ t_d $, and $ r_d $ are the scalar entries.
\end{itemize}

TransE \cite{bordes_translatingembeddingsmodeling_2013} was the first model of this type, with score function basically the same as the above equation. There have been many extensions such as TransR \cite{lin_learningentityrelation_2015}, TransH \cite{wang_knowledgegraphembedding_2014}, and TransA \cite{xiao_transaadaptiveapproach_2015}. Most extensions are done by linear transformation of the entities into a relation-specific space before translation \cite{lin_learningentityrelation_2015}.

These models are simple and efficient. However, their modeling capacity is generally weak because of over-strong assumptions about translation using relation embedding. Therefore, they are unable to model some forms of data \cite{wang_multirelationallinkprediction_2018}.

\subsubsection{Neural-network-based:} These models use a nonlinear neural network to compute the matching score for a triple:
\begin{equation} \label{eq:nn}
\begin{split}
\gS(h,t,r) = &NN(\vh, \vt, \vr),
\end{split}
\end{equation}
where 
\begin{itemize}
	\item $ \vh, \vt, \vr $ are the embedding vectors of $ h $, $ t $, and $ r $, respectively,
	\item $ NN $ is the neural network used to compute the score.
\end{itemize}

One of the simplest neural-network-based model is ER-MLP \cite{dong_knowledgevaultwebscale_2014}, which concatenates the input embedding vectors and uses a multi-layer perceptron neural network to compute the matching score.
NTN \cite{socher_reasoningneuraltensor_2013} is an earlier model that employs nonlinear activation functions to generalize the linear model RESCAL \cite{nickel_threewaymodelcollective_2011}. Recent models such as ConvE \cite{dettmers_convolutional2dknowledge_2018} use convolution networks instead of fully-connected networks.

These models are complicated because of their use of neural networks as a black-box universal approximator, which usually make them difficult to understand and expensive to use.

\subsubsection{Trilinear-product-based:} These models compute their scores by using trilinear product between head, tail, and relation embeddings, with relation embedding playing the role of matching weights on the dimensions of head and tail embeddings:
\begin{equation} \label{eq:trilinear}
\begin{split}
\gS(h,t,r) = &\langle \vh, \vt, \vr \rangle \\
= & \vh^\top diag(\vr) \vt \\
= &\sum_{d=1}^{D} \left(\vh \odot \vt \odot \vr\right)_d \\
= &\sum_{d=1}^{D} \left(h_d t_d r_d\right),
\end{split}
\end{equation}
where 
\begin{itemize}
	\item $ \vh, \vt, \vr $ are embedding vectors of $ h $, $ t $, and $ r $, respectively,
	\item $ diag(\vr) $ is the diagonal matrix of $ \vr $,
	\item $ \odot $ denotes the element-wise Hadamard product,
	\item $ D $ is the embedding size and $ d $ is the dimension for which $ h_d $, $ t_d $, and $ r_d $ are the scalar entries.
\end{itemize}

In this paper, we focus on this category, particularly on DistMult, ComplEx, CP, and CP$ _h $ with augmented data. These models are simple, efficient, and can scale linearly with respect to embedding size in both time and space. They are also very effective, as has been shown by the state-of-the-art results for ComplEx and CP$ _h $ using popular benchmarks \cite{trouillon_complexembeddingssimple_2016} \cite{lacroix_canonicaltensordecomposition_2018}.

DistMult \cite{yang_embeddingentitiesrelations_2015} embeds each entity and relation as a single real-valued vector. DistMult is the simplest model in this category. Its score function is symmetric, with the same scores for triples $ (h, t, r) $ and $ (t, h, r) $. Therefore, it cannot model asymmetric data for which only one direction is valid, e.g., asymmetric triples such as \textit{(Paper1, Paper2, cite)}. Its score function is:
\begin{equation} \label{eq:distmult}
\begin{split}
\gS(h,t,r) = &\langle \vh, \vt, \vr \rangle,
\end{split}
\end{equation}
where $ \vh, \vt, \vr \in \sR^{D} $.

ComplEx \cite{trouillon_complexembeddingssimple_2016} is an extension of DistMult that uses complex-valued embedding vectors that contain complex numbers. Each complex number $ c $ with two components, real $ a $ and imaginary $ b $, can be denoted as $ c = a + b i $. The complex conjugate $ \overline{c} $ of $ c $ is $ \overline{c} = a - b i $. The complex conjugate vector $ \overline{\vt} $ of $ \vt $ is form from the complex conjugate of the individual entries. Complex algebra requires using the complex conjugate vector of tail embedding in the inner product and trilinear product \cite{ahlfors_complexanalysisintroduction_1953}. Thus, these products can be antisymmetric, which enables ComplEx to model asymmetric data \cite{trouillon_complexembeddingssimple_2016} \cite{trouillon_knowledgegraphcompletion_2017}. Its score function is:
\begin{equation} \label{eq:complex}
\begin{split}
\gS(h,t,r) = &\text{Re}(\langle \vh, \overline{\vt}, \vr \rangle),
\end{split}
\end{equation}
where $ \vh, \vt, \vr \in \sC^{D} $ and Re$ (c) $ means taking the real component of the complex number $ c $.

CP \cite{hitchcock_expressiontensorpolyadic_1927} is similar to DistMult but embeds entities as head and as tail differently. Each entity $ e $ has two embedding vectors $ \ve $ and $ \ve^{(2)} $ depending on its role in a triple as head or as tail, respectively. Using different role-based embedding vectors leads to an asymmetric score function, enabling CP to also model asymmetric data. However, experiments have shown that CP's performance is very poor on unseen test data \cite{lacroix_canonicaltensordecomposition_2018}. Its score function is:
\begin{equation} \label{eq:cp}
\begin{split}
\gS(h,t,r) = &\langle \vh, \vt^{(2)}, \vr \rangle,
\end{split}
\end{equation}
where $ \vh, \vt^{(2)}, \vr \in \sR^{D} $.

CP$ _h $ \cite{lacroix_canonicaltensordecomposition_2018} is a direct extension of CP. Its heuristic augments the training data by making an inverse triple $ (t, h, r^{(a)}) $ for each existing triple $ (h, t, r) $, where $ r^{(a)} $ is the augmented relation corresponding to $ r $. With this heuristic, CP$ _h $ significantly improves CP, achieving results competitive with ComplEx. Its score function is:
\begin{equation} \label{eq:cph}
\begin{split}
\gS(h,t,r) = &\langle \vh, \vt^{(2)}, \vr \rangle \text{ and } \langle \vt, \vh^{(2)}, \vr^{(a)} \rangle,
\end{split}
\end{equation}
where $ \vh, \vh^{(2)}, \vt, \vt^{(2)}, \vr, \vr^{(a)} \in \sR^{D} $.

In the next section, we present a new approach to analyzing these trilinear-product-based models.

\section{Multi-Embedding Interaction} \label{sect:method}
In this section, we first formally present the multi-embedding interaction mechanism. We then derive each of the above trilinear-product-based models using this mechanism, by changing the embedding vectors and setting appropriate weight vectors. Next, we specify our attempt at learning weight vectors automatically. We also propose a four-embedding interaction model based on quaternion algebra.

\subsection{Multi-embedding interaction mechanism}
We globally model each entity $ e $ as the multiple embedding vectors $ \{\ve^{(1)}, \ve^{(2)}, \dots, \ve^{(n)}\} $ and each relation $ r $ as the multiple embedding vectors $ \{\vr^{(1)}, \vr^{(2)}, \dots, \vr^{(n)}\} $, where $ n $ is the number of multi-embedding vectors generally, which can be different for entity and relation. The triple $ (h, t, r) $ is therefore modeled by multi-embedding vectors $ \vh^{(i)}, \vt^{(j)}, \text{ and } \vr^{(k)}, i, j, k \in \{1, \dots, n\} $.

In each triple, the embedding vectors for head, tail, and relation interact with each and every other embedding vector to produce multiple interaction scores. Each interaction is modeled by the trilinear product of corresponding embedding vectors. The interaction scores are then weighted summed by a weight vector:
\begin{equation} \label{eq:score}
\begin{split}
\gS (h,t,r;\bm{\Theta}, \bm{\omega}) &= \sum_{i, j, k \in \{1, ..., n\}} \omega^{(i,j,k)} \langle \vh^{(i)}, \vt^{(j)}, \vr^{(k)} \rangle,
\end{split}
\end{equation}
where 
\begin{itemize}
	\item $ \bm{\omega} \in \sR^{n^3} $ denotes the weight vector used to combine the interaction scores, with $ \omega^{(i,j,k)} $ being an element,
	\item $ \bm{\Theta} $ denotes embedding vectors $ \vh^{(i)}, \vt^{(j)}, \vr^{(k)} $.
\end{itemize}

\subsection{Deriving trilinear-product-based models}
\begin{table*}[ht]
%	\small
	\centering
	
	\caption{Weight vectors for special cases.}
	\label{tab:weight_vector}
	
	\begin{tabular}{c|c|c|c|c|c|c|c|c}
		\hline 
		\textbf{Weighted terms} & \textbf{DistMult} & \textbf{ComplEx} & \makecell{\textbf{ComplEx} \\\textbf{equiv. 1}} & \makecell{\textbf{ComplEx} \\\textbf{equiv. 2}} & \makecell{\textbf{ComplEx} \\\textbf{equiv. 3}} & \textbf{CP} & \textbf{CP$ _h $} & \makecell{\textbf{CP$ _h $} \\\textbf{equiv.}}\\ 
		\hline
		
		$ \langle \vh^{(1)}, \vt^{(1)}, \vr^{(1)} \rangle $ & 1 & 1 & 1 & 0 & 0 & 0 & 0 & 0\\ 
		$ \langle \vh^{(1)}, \vt^{(1)}, \vr^{(2)} \rangle $ & 0 & 0 & 0 & 1 & 1 & 0 & 0 & 0\\ 
		$ \langle \vh^{(1)}, \vt^{(2)}, \vr^{(1)} \rangle $ & 0 & 0 & 0 & -1 & 1 & 1 & 1 & 0\\ 
		$ \langle \vh^{(1)}, \vt^{(2)}, \vr^{(2)} \rangle $ & 0 & 1 & -1 & 0 & 0 & 0 & 0 & 1\\ 
		$ \langle \vh^{(2)}, \vt^{(1)}, \vr^{(1)} \rangle $ & 0 & 0 & 0 & 1 & -1 & 0 & 0 & 1\\ 
		$ \langle \vh^{(2)}, \vt^{(1)}, \vr^{(2)} \rangle $ & 0 & -1 & 1 & 0 & 0 & 0 & 1 & 0\\ 
		$ \langle \vh^{(2)}, \vt^{(2)}, \vr^{(1)} \rangle $ & 0 & 1 & 1 & 0 & 0 & 0 & 0 & 0\\ 
		$ \langle \vh^{(2)}, \vt^{(2)}, \vr^{(2)} \rangle $ & 0 & 0 & 0 & 1 & 1 & 0 & 0 & 0\\ 
		\hline 
	\end{tabular}
\end{table*}

The existing trilinear-product-based models can be derived from the proposed general multi-embedding interaction score function in Eq. (\ref{eq:score}) by setting the weight vector $ \bm{\omega} $ as shown in Table \ref{tab:weight_vector}. 

For DistMult, we can see the equivalence directly. For ComplEx, we need to expand its score function following complex algebra \cite{ahlfors_complexanalysisintroduction_1953}:
\begin{equation} \label{eq:complex_expand}
\begin{split}
\gS(h,t,r) = &\text{Re}(\langle \vh, \overline{\vt}, \vr \rangle)\\
= &\langle \text{Re}(\vh), \text{Re}(\vt), \text{Re}(\vr) \rangle + \langle \text{Re}(\vh), \text{Im}(\vt), \text{Im}(\vr) \rangle\\
&- \langle \text{Im}(\vh), \text{Re}(\vt), \text{Im}(\vr) \rangle + \langle \text{Im}(\vh), \text{Im}(\vt), \text{Re}(\vr) \rangle,
\end{split}
\end{equation}
where 
\begin{itemize}
	\item $ \vh, \vt, \vr \in \sC^{D} $,
	\item Re$ (\vc) $ and Im$ (\vc) $ mean taking the real and imaginary components of the complex vector $ \vc $, respectively.
\end{itemize}

Changing Re$ (\vh) $ to $ \vh^{(1)} $, Im$ (\vh) $ to $ \vh^{(2)} $, Re$ (\vt) $ to $ \vt^{(1)} $, Im$ (\vt) $ to $ \vt^{(2)} $, Re$ (\vr) $ to $ \vr^{(1)} $, and Im$ (\vr) $ to $ \vr^{(2)} $, we can rewrite the score function of ComplEx as:
\begin{equation} \label{eq:complex_expand_multiemb}
\begin{split}
\gS(h,t,r) = &\text{Re}(\langle \vh, \overline{\vt}, \vr \rangle)\\
= &\langle \vh^{(1)}, \vt^{(1)}, \vr^{(1)} \rangle + \langle \vh^{(1)}, \vt^{(2)}, \vr^{(2)} \rangle\\
&- \langle \vh^{(2)}, \vt^{(1)}, \vr^{(2)} \rangle + \langle \vh^{(2)}, \vt^{(2)}, \vr^{(1)} \rangle,
\end{split}
\end{equation}
which is equivalent to the weighted sum using the weight vectors in Table \ref{tab:weight_vector}. Note that by the symmetry between $ h $ and $ t $, we can also obtain the equivalent weight vector \textbf{ComplEx} \textbf{equiv. 1}. By symmetry between embedding vectors of the same entity or relation, we can also obtain the equivalent weight vectors \textbf{ComplEx} \textbf{equiv. 2} and \textbf{ComplEx} \textbf{equiv. 3}.

For CP, note that the two role-based embedding vectors for each entity can be mapped to two-embedding vectors in our model and the relation embedding vector can be mapped to $ \vr^{(1)} $. For CP$ _h $, further note that its data augmentation is equivalent to adding the score of the original triple and the inverse triple when training using stochastic gradient descent (SGD):
\begin{equation} \label{eq:cph_add}
\begin{split}
\gS(h,t,r) = &\langle \vh, \vt^{(2)}, \vr \rangle + \langle \vt, \vh^{(2)}, \vr^{(a)} \rangle.
\end{split}
\end{equation}
We can then map $ \vr^{(a)} $ to $ \vr^{(2)} $ to obtain the equivalence given in Table \ref{tab:weight_vector}. By symmetry between $ h $ and $ t $, we can also obtain the equivalent weight vector \textbf{CP$ _h $} \textbf{equiv. 1}.

From this perspective, all four models DistMult, ComplEx, CP, and CP$ _h $ can be seen as special cases of the general multi-embedding interaction mechanism. This provides an intuitive perspective on using the embedding vectors in complicated models. For the ComplEx model, instead of using a complex-valued embedding vector, we can treat it as two real-valued embedding vectors. These vectors can then be used directly in common learning algorithms that take as input real-valued vectors rather than complex-valued vectors. We also see that multiple embedding vectors are a natural extension of single embedding vectors. Given this insight, multiple embedding vectors can be concatenated to form a longer vector for use in visualization and data analysis, for example.

%We analyze the performance of the above derived weight vectors and other variants in Section \ref{sect:result}.

\subsection{Automatically learning weight vectors}
As we have noted, the weight vector $ \bm{\omega} $ plays an important role in the model, because it determines how the interaction mechanism is implemented and therefore how the specific model can be derived. An interesting question is how to learn $ \bm{\omega} $ automatically. One approach is to let the model learn $ \bm{\omega} $ together with the embeddings in an end-to-end fashion. For a more detailed examination of this idea, we will test different restrictions on the range of $ \bm{\omega} $ by applying $ \text{tanh}(\bm{\omega}) $, $ \text{sigmoid}(\bm{\omega}) $, and $ \text{softmax}(\bm{\omega}) $.

Note also that the weight vectors for related models are usually sparse. We therefore enforce a sparsity constraint on $ \bm{\omega} $ by an additional Dirichlet negative log-likelihood regularization loss:
\begin{equation} \label{eq:dirichlet}
\begin{split}
\gL_{dir} = - \lambda_{dir} \sum_{i, j, k \in \{1, ..., n\}} (\alpha - 1) \log \frac{|\omega^{(i, j, k)}|}{||\bm{\omega}||_1},
\end{split}
\end{equation}
where $ \alpha $ is a hyperparameter controlling sparseness (the smaller $ \alpha $ is, the sparser $ \bm{\omega} $ is) and $ \lambda_{dir} $ is the regularization strength.

%We evaluate these settings in Section \ref{sect:result}.

\subsection{Quaternion-based four-embedding interaction model}
Another question is whether using more embedding vectors in the multi-embedding interaction mechanism is helpful. Motivated by the derivation of ComplEx from a two-embedding interaction model, we develop a four-embedding interaction model by using quaternion algebra to determine the weight vector and the interaction mechanism.

Quaternion numbers are extension of complex numbers to four components \cite{kantor_hypercomplexnumberselementary_1989} \cite{goldman_rethinkingquaternions_2010}. Each quaternion number $ q $, with one real component $ a $ and three imaginary components $ b, c, d $, could be written as $ q = a + b \vi + c \vj + d \vk $ where $ \vi, \vj, \vk $ are fundamental quaternion units, similar to the imaginary number $ i $ in complex algebra. As for complex conjugates, we also have a quaternion conjugate $ \overline{q} = a - b \vi - c \vj - d \vk $.

An intuitive view of quaternion algebra is that each quaternion number represents a 4-dimensional vector (or 3-dimensional vector when the real component $ a = 0 $) and quaternion multiplication is rotation of this vector in 4- (or 3-)dimensional space. Compared to complex algebra, each complex number represents a 2-dimensional vector and complex multiplication is rotation of this vector in 2-dimensional plane \cite{ahlfors_complexanalysisintroduction_1953}.

Several works have shown the benefit of using complex, quaternion, or other hyper-complex numbers in the hidden layers of deep neural networks \cite{guberman_complexvaluedconvolutional_2016} \cite{minemoto_feedforwardneural_2017} \cite{parcollet_quaternionrecurrentneural_2019}. To the best of our knowledge, this paper is the first to motivate and use quaternion numbers for the embedding vectors of knowledge graph embedding.

Quaternion multiplication is noncommutative, thus there are multiple ways to multiply three quaternion numbers in the trilinear product. Here, we choose to write the score function of the quaternion-based four-embedding interaction model as:
\begin{equation} \label{eq:quaternion}
\begin{split}
\gS(h,t,r) = &\text{Re}(\langle \vh, \overline{\vt}, \vr \rangle),
\end{split}
\end{equation}
where $ \vh, \vt, \vr \in \sH^{D} $.

By expanding this formula using quaternion algebra \cite{kantor_hypercomplexnumberselementary_1989} and mapping the four components of a quaternion number to four embeddings in the multi-embedding interaction model, respectively, we can write the score function in the notation of the multi-embedding interaction model as:
\begin{equation} \label{eq:quaternion_expand}
\begin{split}
\gS(h,t,r) = &\text{Re}(\langle \vh, \overline{\vt}, \vr \rangle)\\
= &\langle \vh^{(1)}, \vt^{(1)}, \vr^{(1)} \rangle + \langle \vh^{(2)}, \vt^{(2)}, \vr^{(1)} \rangle\\
&+ \langle \vh^{(3)}, \vt^{(3)}, \vr^{(1)} \rangle + \langle \vh^{(4)}, \vt^{(4)}, \vr^{(1)} \rangle\\
&+ \langle \vh^{(1)}, \vt^{(2)}, \vr^{(2)} \rangle - \langle \vh^{(2)}, \vt^{(1)}, \vr^{(2)} \rangle\\
&+ \langle \vh^{(3)}, \vt^{(4)}, \vr^{(2)} \rangle - \langle \vh^{(4)}, \vt^{(3)}, \vr^{(2)} \rangle\\
&+ \langle \vh^{(1)}, \vt^{(3)}, \vr^{(3)} \rangle - \langle \vh^{(2)}, \vt^{(4)}, \vr^{(3)} \rangle\\
&- \langle \vh^{(3)}, \vt^{(1)}, \vr^{(3)} \rangle + \langle \vh^{(4)}, \vt^{(2)}, \vr^{(3)} \rangle\\
&+ \langle \vh^{(1)}, \vt^{(4)}, \vr^{(4)} \rangle + \langle \vh^{(2)}, \vt^{(3)}, \vr^{(4)} \rangle\\
&- \langle \vh^{(3)}, \vt^{(2)}, \vr^{(4)} \rangle - \langle \vh^{(4)}, \vt^{(1)}, \vr^{(4)} \rangle,
\end{split}
\end{equation}
where $ \vh, \vt, \vr \in \sH^{D} $.

%We evaluate this quaternion-based four-embedding interaction model in Section \ref{sect:result}.

\section{Loss Function and Optimization}
The learning problem in knowledge graph embedding methods can be modeled as the binary classification of valid and invalid triples. Because knowledge graphs do not contain invalid triples, we generate them by negative sampling \cite{mikolov_efficientestimationword_2013}. For each valid triple $ (h, t, r) $, we replace the $ h $ or $ t $ entities in each training triple with other random entities to obtain the invalid triples $ (h', t, r) $ and $ (h, t', r) $ \cite{bordes_translatingembeddingsmodeling_2013}. 

We can then learn the model parameters by minimizing the negative log-likelihood loss for the training data with the predicted probability modeled by the logistic sigmoid function $ \sigma (\cdot) $ on the matching score. This loss is the cross-entropy:
\begin{equation} \label{eq:loss_crossentropy} % fixed typo in loss function
\begin{split}
\gL_1(\gD, \gD';\bm{\Theta}, \bm{\omega}) = &- \sum_{(h, t, r) \in \gD}{\log \sigma \left( \gS \left( h,t,r;\bm{\Theta}, \bm{\omega} \right) \right)} \\
& - \sum_{(h', t', r) \in \gD'}{\log \left( 1 - \sigma \left( \gS \left( h',t',r;\bm{\Theta}, \bm{\omega} \right) \right) \right)},
\end{split}
\end{equation}
where $ \gD $ is true data ($ \hat{p} = 1 $), $ \gD' $ is negative sampled data ($ \hat{p} = 0 $), and $ \hat{p} $ is the empirical probability.

Defining the class label $ \text{Y}_{(h,t,r)} = 2 \hat{p}_{(h,t,r)} - 1 $, i.e., the labels of positive triples are $ 1 $ and negative triples are $ -1 $, the above loss can be written more concisely as:
\begin{equation} \label{eq:loss_softplus} % used consistent size bracket
\begin{split}
\gL_2(\gD, \gD';\bm{\Theta}, \bm{\omega}) = \sum_{(h, t, r) \in \gD \cup \gD'}&\log \bigl( 1 + \text{e}^{- \text{Y}_{(h,t,r)} \gS (h,t,r;\bm{\Theta}, \bm{\omega})} \bigr),
\end{split}
\end{equation}
where $ \gD $ is true data and $ \gD' $ is negative sampled data. The equivalence of these losses are shown in the appendix.

Including the regularization, such as by $ L^2 $ loss on the embedding vectors, it can be written as:
\begin{equation} \label{eq:loss_softplus_reg}
	\begin{split}
		\gL(\gD, \gD';\bm{\Theta}, \bm{\omega}) = \sum_{(h, t, r) \in \gD \cup \gD'}&\Bigl(\log \bigl( 1 + \text{e}^{- \text{Y}_{(h,t,r)} \gS (h,t,r;\bm{\Theta}, \bm{\omega})} \bigr) \\
		&+ \frac{\lambda}{n D} ||\bm{\Theta}||_2^2 \Bigr),
	\end{split}
\end{equation}
where $ \bm{\Theta} $ are the embedding vectors of specific triple $ (h, t, r) $, $ n D $ is the total embedding size, and $ \lambda $ is the regularization strength.

\section{Experimental Settings}
\subsection{Datasets}
For our empirical analysis, we used the WN18 dataset, the most popular of the benchmark datasets built on WordNet \cite{millergeorgea._wordnetlexicaldatabase_1995} by Bordes et al. \cite{bordes_translatingembeddingsmodeling_2013}. This dataset has 40,943 \textit{entities}, 18 \textit{relations}, 141,442 \textit{training triples}, 5,000 \textit{validation triples}, 5,000 \textit{test triples}. In our preliminary experiments, the relative performance on all datasets was quite consistent, therefore choosing the WN18 dataset is appropriate for our analysis. We will consider the use of other datasets in in future work.

\subsection{Evaluation protocols}
Knowledge graph embedding methods are usually evaluated on link prediction task \cite{bordes_translatingembeddingsmodeling_2013}. In this task, for each true triple $ (h, t, r) $ in the test set, we replace $ h $ and $ t $ by every other entity to generate corrupted triples $ (h', t, r) $ and $ (h, t', r) $, respectively \cite{bordes_translatingembeddingsmodeling_2013}. The goal of the model now is to rank the true triple $ (h, t, r) $ before the corrupted triples based on the predicted score $ \gS $.

For each true triple in the test set, we compute its rank, then we can compute popular evaluation metrics including $ MRR $ (mean reciprocal rank) and $ Hit@k $ for $ k \in \{1, 3, 10\} $ (how many true triples are correctly ranked in the top $ k $) \cite{trouillon_complexembeddingssimple_2016}. 

To avoid false negative error, i.e., corrupted triples are accidentally valid triples, we follow the protocols used in other works for filtered metrics \cite{bordes_translatingembeddingsmodeling_2013}. In this protocol, all valid triples in the training, validation, and test sets are removed from the corrupted triples set before computing the rank of the true triple.

\subsection{Training}
We trained the models using SGD with learning rates auto-tuned by Adam \cite{kingma_adammethodstochastic_2014}, that makes the choice of initial learning rate more robust. For all models, we found good hyperparameters with grid search on learning rates $ \in \{10^{-3}, 10^{-4}\} $, embedding regularization strengths $ \in \{10^{-2}, 3 \times 10^{-3}, 10^{-3}, 3 \times 10^{-4}, 10^{-4}, 0.0\} $, and batch sizes $ \in \{2^{12}, 2^{14}\} $. For a fair comparison, we fixed the embedding sizes so that numbers of parameters for all models are comparable. In particular, we use embedding sizes of $ 400 $ for one-embedding models such as DistMult, $ 200 $ for two-embedding models such as ComplEx, CP, and CP$ _h $, and $ 100 $ for four-embedding models. We also fixed the number of negative samples at $ 1 $ because, although using more negative samples is beneficial for all models, it is also more expensive and not necessary for this comparative analysis.

We constrained entity embedding vectors to have unit \textit{$ L^2 $-norm} after each training iteration. All training runs were stopped early by checking the filtered MRR on the validation set after every 50 epochs, with 100 epochs patient.

\section{Results and Discussion} \label{sect:result}
In this section, we present experimental results and analyses for the models described in Section \ref{sect:method}. We report results for derived weight vectors and their variants, auto-learned weight vectors, and the quaternion-based four-embedding interaction model.

\begin{table*}[ht]
%	\small
	\centering
	
	\caption{Results for the derived weight vectors on WN18.}
	\label{tab:result}
	
	\begin{tabular}{l|c|c|c|c}
		\hline 
		\textbf{Weight setting} & \textbf{MRR} & \textbf{Hit@1} & \textbf{Hit@3} & \textbf{Hit@10}\\ 
		\hline
		
		DistMult $ (1, 0, 0, 0, 0, 0, 0, 0) $ & 0.796 & 0.674 & 0.915 & 0.945\\ 
		ComplEx $ (1, 0, 0, 1, 0, -1, 1, 0) $ & \textbf{0.937} & \textbf{0.928} & \textbf{0.946} & \textbf{0.951}\\ 
		CP $ (0, 0, 1, 0, 0, 0, 0, 0) $ & 0.086 & 0.059 & 0.093 & 0.139\\ 
		CP$ _h $ $ (0, 0, 1, 0, 0, 1, 0, 0) $ & \textbf{0.937} & \textbf{0.929} & \textbf{0.944} & \textbf{0.949}\\ 
		\hline
		
		DistMult \textit{on train} & 0.917 & 0.848 & 0.985 & 0.997\\ 
		ComplEx \textit{on train} & \textit{0.996} & \textit{0.994} & \textit{0.998} & \textit{0.999}\\ 
		CP \textit{on train} & \textit{0.994} & \textit{0.994} & \textit{0.996} & \textit{0.999}\\ 
		CP$ _h $ \textit{on train} & \textit{0.995} & \textit{0.994} & \textit{0.998} & \textit{0.999}\\ 
		\hline
		
		Bad example 1 $ (0, 0, 20, 0, 0, 1, 0, 0) $ & 0.107 & 0.079 & 0.116 & 0.159\\ 
		Bad example 2 $ (0, 0, 1, 1, 1, 1, 0, 0) $ & 0.794 & 0.666 & 0.917 & 0.947\\ 
		
		Good example 1 $ (0, 0, 20, 1, 1, 20, 0, 0) $ & \textbf{0.938} & \textbf{0.934} & \textbf{0.942} & \textbf{0.946}\\
		Good example 2 $ (1, 1, -1, 1, 1, -1, 1, 1) $ & \textbf{0.938} & \textbf{0.930} & \textbf{0.944} & \textbf{0.950}\\ 
		\hline 
	\end{tabular}
\end{table*}

\begin{table*}[ht]
%	\small
	\centering
	
	\caption{Results for the auto-learned weight vectors on WN18.}
	\label{tab:result_learnweight}
	
	\begin{tabular}{l|c|c|c|c}
		\hline 
		\textbf{Weight setting} & \textbf{MRR} & \textbf{Hit@1} & \textbf{Hit@3} & \textbf{Hit@10}\\ 
		\hline
		
		Uniform weight $ (1, 1, 1, 1, 1, 1, 1, 1) $ & 0.787 & 0.658 & 0.915 & 0.944\\ 
		\hline
		
		Auto weight no restriction & 0.774 & 0.636 & 0.911 & 0.944\\ 
		Auto weight $ \in (-1, 1) $ by tanh & 0.765 & 0.625 & 0.908 & 0.943\\ 
		Auto weight $ \in (0, 1) $ by sigmoid & 0.789 & 0.661 & 0.915 & 0.946\\ 
		Auto weight $ \in (0, 1) $ by softmax & 0.802 & 0.685 & 0.915 & 0.944\\ 
		Auto weight no restriction, sparse & 0.792 & 0.685 & 0.892 & 0.935\\ 
		Auto weight $ \in (-1, 1) $ by tanh, sparse & 0.763 & 0.613 & 0.910 & 0.943\\ 
		Auto weight $ \in (0, 1) $ by sigmoid, sparse & 0.793 & 0.667 & 0.915 & 0.945\\ 
		Auto weight $ \in (0, 1) $ by softmax, sparse & 0.803 & 0.688 & 0.915 & 0.944\\ 
		\hline 
	\end{tabular}
\end{table*}

\begin{table*}[ht]
%	\small
	\centering
	
	\caption{Results for the quaternion-based four-embedding interaction model on WN18.}
	\label{tab:result_quaternion}
	
	\begin{tabular}{l|c|c|c|c}
		\hline 
		\textbf{Weight setting} & \textbf{MRR} & \textbf{Hit@1} & \textbf{Hit@3} & \textbf{Hit@10}\\ 
		\hline
		
		Quaternion-based four-embedding & \textbf{0.941} & \textbf{0.931} & \textbf{0.950} & \textbf{0.956}\\ 
		Quaternion-based four-embedding \textit{on train} & \textit{0.997} & \textit{0.995} & \textit{0.999} & \textit{1.000}\\ 
		\hline 
	\end{tabular}
\end{table*}

\subsection{Derived weight vectors and variants}
\subsubsection{Comparison of derived weight vectors } 
We evaluated the multi-embedding interaction model with the score function in Eq. (\ref{eq:score}), using the derived weight vectors in Table \ref{tab:weight_vector}. The results are shown in Table \ref{tab:result}. They are consistent with the results reported in other works \cite{trouillon_complexembeddingssimple_2016}. Note that ComplEx and CP$ _h $ achieved good results, whereas DistMult performed less well. CP performed very poorly in comparison to the other models, even though it is a classical model for the tensor decomposition task \cite{hitchcock_expressiontensorpolyadic_1927}.

For a more detailed comparison, we report the performance on training data. Note that ComplEx and CP$ _h $ can accurately predict the training data, whereas DistMult did not. This is evidence that ComplEx and CP$ _h $ are fully expressive while DistMult cannot model asymmetric data effectively. 

The most surprising result was that CP can also accurately predict the training data at a comparable level to ComplEx and CP$ _h $, despite its very poor result on the test data. This suggests that the problem with CP is not its modeling capacity, but in its generalization performance to new test data. In other words, CP is severely overfitting to the training data. However, standard regularization techniques such as $ L^2 $ regularization did not appear to help. CP$ _h $ can be seen as a regularization technique that does help CP generalize well to unseen data.

\subsubsection{Comparison with other variants of weight vectors} 
In Table \ref{tab:result}, we show the results for two bad examples and two good examples of weight vector variants. Note that \textit{bad example 1} performed similarly to CP and \textit{bad example 2} performed similarly to DistMult. \textit{Good example 1} was similar to CP$ _h $ and \textit{good example 2} was similar to ComplEx.

This shows that the problem of bad weight vectors is not unique to some specific models. Moreover, it shows that there are other good weight vectors, besides those for ComplEx and CP$ _h $, that can achieve very good results.

We note that the good weight vectors exhibit the following properties.
\begin{itemize}
	\item \textit{Completeness}: all embedding vectors in a triple should be involved in the weighted-sum matching score.
	\item \textit{Stability}: all embedding vectors for the same entity or relation should contribute equally to the weighted-sum matching score.
	\item \textit{Distinguishability}: the weighted-sum matching scores for different triples should be \textit{distinguishable}. For example, the score $ \langle \vh^{(1)}, \vt^{(2)}, \vr^{(1)} \rangle + \langle \vh^{(2)}, \vt^{(1)}, \vr^{(2)} \rangle $ is \textit{indistinguishable} because switching $ h $ and $ t $ forms a symmetric group.
\end{itemize}

As an example, consider the ComplEx model, where the multiplication of two complex numbers written in polar coordinate format, $ c_1 = |c_1|\text{e}^{-\text{i}\theta_1} $ and $ c_2 = |c_2|\text{e}^{-\text{i}\theta_2} $, can be written as $ c_1 c_2 = |c_1||c_2|\text{e}^{-\text{i}(\theta_1 + \theta_2)} $ \cite{ahlfors_complexanalysisintroduction_1953}. This is a rotation in the complex plane, which intuitively satisfies the above properties.

\subsection{Automatically learned weight vectors} 
We let the models learn $ \bm{\omega} $ together with the embeddings in an end-to-end fashion, aiming to learn good weight vectors automatically. The results are shown in Table \ref{tab:result_learnweight}.

We first set uniform weight vector as a baseline. The results were similar to those for DistMult because the weighted-sum matching score is also symmetric. However, other automatically learned weight vectors also performed similarly to DistMult. Different restrictions by applying $ \text{tanh}(\bm{\omega}) $, $ \text{sigmoid}(\bm{\omega}) $, and $ \text{softmax}(\bm{\omega}) $ did not help. We noticed that the learned weight vectors were almost uniform, making them \textit{indistinguishable}, suggesting that the use of sparse weight vectors might help.

We enforced a sparsity constraint by an additional Dirichlet negative log-likelihood regularization loss on $ \bm{\omega} $, with $ \alpha $ tuned to $ \frac{1}{16} $ and $ \lambda_{dir} $ tuned to $ 10^{-2} $. However, the results did not improve. Tracking of weight vectors value showed that the sparsity constraint seemed to amplify the initial differences between the weight values instead of learning useful sparseness. This suggests that the gradient information is too symmetric that the model cannot break the symmetry of $ \bm{\omega} $ and escape the local optima \footnote{Note that we also tried $ L^1 $ regularization to enforce sparsity but it did not appear to help. Theoretically, $ L^0 $ regularization would successfully find the optimal $ \bm{\omega} $ but solving it exactly is NP-hard.}.

In general, these experiments show that learning good weight vectors automatically is a particularly difficult task.

\subsection{Quaternion-based four-embedding interaction model} 
In Table \ref{tab:result_quaternion}, we present the evaluation results for the proposed quaternion-based four-embedding interaction model. The results were generally positive, with most metrics higher than those in Table \ref{tab:result} for state-of-the-art models such as ComplEx and CP$ _h $. Especially, H@10 performance was much better than other models.

Note that this model needs more extensive evaluation. One potential problem is its being prone to overfitting, as seen in the \textit{on train} results, with H@10 at absolute 1.000. This might mean that better regularization methods may be needed. However, the general results suggest that extending to more embedding vectors for multi-embedding interaction models is a promising approach.

\section{Conclusion}
This paper proposes a multi-embedding interaction mechanism as a new approach to analyzing state-of-the-art knowledge graph embedding models such as DistMult, ComplEx, CP, and CP$ _h $. We show that these models can be unified and generalized under the new approach to provide an intuitive perspective on using the models and their embedding vectors effectively. We analyzed and compared the models and their variants empirically to better understand their properties, such as the severe overfitting problem of the CP model. In addition, we propose and have evaluated a new multi-embedding interaction model based on quaternion algebra, which showed some promising results.

There are several promising future directions. One direction is to find new methods of modeling the interaction mechanism between multi-embedding vectors and the effective extension to additional embedding vectors. Another direction is to evaluate multi-embedding models such as the proposed quaternion-based four-embedding interaction model more extensively.

\begin{acks}
This work was supported by a JSPS Grant-in-Aid for Scientific Research (B) (15H02789).
\end{acks}

\bibliographystyle{ACM-Reference-Format}
% References section
%\bibliography{../../../Zotero.bib}

\begin{thebibliography}{36}

%%% the Web address.

\ifx \showCODEN    \undefined \def \showCODEN     #1{\unskip}     \fi
\ifx \showDOI      \undefined \def \showDOI       #1{#1}\fi
\ifx \showISBNx    \undefined \def \showISBNx     #1{\unskip}     \fi
\ifx \showISBNxiii \undefined \def \showISBNxiii  #1{\unskip}     \fi
\ifx \showISSN     \undefined \def \showISSN      #1{\unskip}     \fi
\ifx \showLCCN     \undefined \def \showLCCN      #1{\unskip}     \fi
\ifx \shownote     \undefined \def \shownote      #1{#1}          \fi
\ifx \showarticletitle \undefined \def \showarticletitle #1{#1}   \fi
\ifx \showURL      \undefined \def \showURL       {\relax}        \fi
% invisible to TeX
\providecommand\bibfield[2]{#2}
\providecommand\bibinfo[2]{#2}
\providecommand\natexlab[1]{#1}
\providecommand\showeprint[2][]{arXiv:#2}

\bibitem[\protect\citeauthoryear{Ahlfors}{Ahlfors}{1953}]%
        {ahlfors_complexanalysisintroduction_1953}
\bibfield{author}{\bibinfo{person}{Lars~V. Ahlfors}.}
  \bibinfo{year}{1953}\natexlab{}.
\newblock \showarticletitle{Complex Analysis: An Introduction to the Theory of
  Analytic Functions of One Complex Variable}.
\newblock \bibinfo{journal}{\emph{New York, London}} (\bibinfo{year}{1953}),
  \bibinfo{pages}{177}.
\newblock


\bibitem[\protect\citeauthoryear{{Amit Singhal}}{{Amit Singhal}}{2012}]%
        {amitsinghal_officialgoogleblog_2012}
\bibfield{author}{\bibinfo{person}{{Amit Singhal}}.}
  \bibinfo{year}{2012}\natexlab{}.
\newblock \bibinfo{title}{Official {{Google Blog}}: {{Introducing}} the
  {{Knowledge Graph}}: Things, Not Strings}.
\newblock
  \bibinfo{howpublished}{https://googleblog.blogspot.com/2012/05/introducing-knowledge-graph-things-not.html}.
\newblock


\bibitem[\protect\citeauthoryear{Bollacker, Evans, Paritosh, Sturge, and
  Taylor}{Bollacker et~al\mbox{.}}{2008}]%
        {bollacker_freebasecollaborativelycreated_2008}
\bibfield{author}{\bibinfo{person}{Kurt Bollacker}, \bibinfo{person}{Colin
  Evans}, \bibinfo{person}{Praveen Paritosh}, \bibinfo{person}{Tim Sturge},
  {and} \bibinfo{person}{Jamie Taylor}.} \bibinfo{year}{2008}\natexlab{}.
\newblock \showarticletitle{Freebase: A Collaboratively Created Graph Database
  for Structuring Human Knowledge}. In \bibinfo{booktitle}{\emph{In {{SIGMOD
  Conference}}}}. \bibinfo{pages}{1247--1250}.
\newblock


\bibitem[\protect\citeauthoryear{Bordes, Usunier, {Garcia-Duran}, Weston, and
  Yakhnenko}{Bordes et~al\mbox{.}}{2013}]%
        {bordes_translatingembeddingsmodeling_2013}
\bibfield{author}{\bibinfo{person}{Antoine Bordes}, \bibinfo{person}{Nicolas
  Usunier}, \bibinfo{person}{Alberto {Garcia-Duran}}, \bibinfo{person}{Jason
  Weston}, {and} \bibinfo{person}{Oksana Yakhnenko}.}
  \bibinfo{year}{2013}\natexlab{}.
\newblock \showarticletitle{Translating Embeddings for Modeling
  Multi-Relational Data}. In \bibinfo{booktitle}{\emph{Advances in {{Neural
  Information Processing Systems}}}}. \bibinfo{pages}{2787--2795}.
\newblock


\bibitem[\protect\citeauthoryear{{Carrer-Neto}, {Hern\'andez-Alcaraz},
  {Valencia-Garc\'ia}, and {Garc\'ia-S\'anchez}}{{Carrer-Neto}
  et~al\mbox{.}}{2012}]%
        {carrer-neto_socialknowledgebasedrecommender_2012}
\bibfield{author}{\bibinfo{person}{Walter {Carrer-Neto}},
  \bibinfo{person}{Mar\'ia~Luisa {Hern\'andez-Alcaraz}},
  \bibinfo{person}{Rafael {Valencia-Garc\'ia}}, {and}
  \bibinfo{person}{Francisco {Garc\'ia-S\'anchez}}.}
  \bibinfo{year}{2012}\natexlab{}.
\newblock \showarticletitle{Social Knowledge-Based Recommender System.
  {{Application}} to the Movies Domain}.
\newblock \bibinfo{journal}{\emph{Expert Systems with Applications}}
  \bibinfo{volume}{39}, \bibinfo{number}{12} (\bibinfo{date}{Sept.}
  \bibinfo{year}{2012}), \bibinfo{pages}{10990--11000}.
\newblock
\showISSN{0957-4174}
\urldef\tempurl%
\url{https://doi.org/10.1016/j.eswa.2012.03.025}
\showDOI{\tempurl}


\bibitem[\protect\citeauthoryear{Dettmers, Minervini, Stenetorp, and
  Riedel}{Dettmers et~al\mbox{.}}{2018}]%
        {dettmers_convolutional2dknowledge_2018}
\bibfield{author}{\bibinfo{person}{Tim Dettmers}, \bibinfo{person}{Pasquale
  Minervini}, \bibinfo{person}{Pontus Stenetorp}, {and}
  \bibinfo{person}{Sebastian Riedel}.} \bibinfo{year}{2018}\natexlab{}.
\newblock \showarticletitle{Convolutional 2d Knowledge Graph Embeddings}. In
  \bibinfo{booktitle}{\emph{In {{Thirty}}-{{Second AAAI Conference}} on
  {{Artificial Intelligence}}}}.
\newblock


\bibitem[\protect\citeauthoryear{Dong, Gabrilovich, Heitz, Horn, Lao, Murphy,
  Strohmann, Sun, and Zhang}{Dong et~al\mbox{.}}{2014}]%
        {dong_knowledgevaultwebscale_2014}
\bibfield{author}{\bibinfo{person}{Xin Dong}, \bibinfo{person}{Evgeniy
  Gabrilovich}, \bibinfo{person}{Geremy Heitz}, \bibinfo{person}{Wilko Horn},
  \bibinfo{person}{Ni Lao}, \bibinfo{person}{Kevin Murphy},
  \bibinfo{person}{Thomas Strohmann}, \bibinfo{person}{Shaohua Sun}, {and}
  \bibinfo{person}{Wei Zhang}.} \bibinfo{year}{2014}\natexlab{}.
\newblock \showarticletitle{Knowledge Vault: A Web-Scale Approach to
  Probabilistic Knowledge Fusion}. In \bibinfo{booktitle}{\emph{Proceedings of
  the 20th {{ACM SIGKDD}} International Conference on {{Knowledge}} Discovery
  and Data Mining - {{KDD}} '14}}. \bibinfo{publisher}{{ACM Press}},
  \bibinfo{address}{New York, New York, USA}, \bibinfo{pages}{601--610}.
\newblock
\showISBNx{978-1-4503-2956-9}
\urldef\tempurl%
\url{https://doi.org/10.1145/2623330.2623623}
\showDOI{\tempurl}


\bibitem[\protect\citeauthoryear{Goldman}{Goldman}{2010}]%
        {goldman_rethinkingquaternions_2010}
\bibfield{author}{\bibinfo{person}{Ron Goldman}.}
  \bibinfo{year}{2010}\natexlab{}.
\newblock \showarticletitle{Rethinking {{Quaternions}}}.
\newblock \bibinfo{journal}{\emph{Synthesis Lectures on Computer Graphics and
  Animation}} \bibinfo{volume}{4}, \bibinfo{number}{1} (\bibinfo{date}{Oct.}
  \bibinfo{year}{2010}), \bibinfo{pages}{1--157}.
\newblock
\showISSN{1933-8996}
\urldef\tempurl%
\url{https://doi.org/10.2200/S00292ED1V01Y201008CGR013}
\showDOI{\tempurl}


\bibitem[\protect\citeauthoryear{Guberman}{Guberman}{2016}]%
        {guberman_complexvaluedconvolutional_2016}
\bibfield{author}{\bibinfo{person}{Nitzan Guberman}.}
  \bibinfo{year}{2016}\natexlab{}.
\newblock \showarticletitle{On {{Complex Valued Convolutional Neural
  Networks}}}.
\newblock \bibinfo{journal}{\emph{arXiv:1602.09046 [cs.NE]}}
  (\bibinfo{date}{Feb.} \bibinfo{year}{2016}).
\newblock
\showeprint[arxiv]{cs.NE/1602.09046}


\bibitem[\protect\citeauthoryear{He, Kang, and McAuley}{He
  et~al\mbox{.}}{2017}]%
        {he_translationbasedrecommendation_2017}
\bibfield{author}{\bibinfo{person}{Ruining He}, \bibinfo{person}{Wang-Cheng
  Kang}, {and} \bibinfo{person}{Julian McAuley}.}
  \bibinfo{year}{2017}\natexlab{}.
\newblock \showarticletitle{Translation-Based {{Recommendation}}}. In
  \bibinfo{booktitle}{\emph{Proceedings of the {{Eleventh ACM Conference}} on
  {{Recommender Systems}}}} \emph{(\bibinfo{series}{{{RecSys}} '17})}.
  \bibinfo{publisher}{{ACM}}, \bibinfo{address}{New York, NY, USA},
  \bibinfo{pages}{161--169}.
\newblock
\showISBNx{978-1-4503-4652-8}
\urldef\tempurl%
\url{https://doi.org/10.1145/3109859.3109882}
\showDOI{\tempurl}


\bibitem[\protect\citeauthoryear{Hinton}{Hinton}{1986}]%
        {hinton_learningdistributedrepresentations_1986}
\bibfield{author}{\bibinfo{person}{Geoffrey~E. Hinton}.}
  \bibinfo{year}{1986}\natexlab{}.
\newblock \showarticletitle{Learning Distributed Representations of Concepts}.
  In \bibinfo{booktitle}{\emph{Proceedings of the Eighth Annual Conference of
  the Cognitive Science Society}}, Vol.~\bibinfo{volume}{1}.
  \bibinfo{publisher}{{Amherst, MA}}, \bibinfo{pages}{12}.
\newblock


\bibitem[\protect\citeauthoryear{Hinton, McClelland, and Rumelhart}{Hinton
  et~al\mbox{.}}{1984}]%
        {hinton_distributedrepresentations_1984}
\bibfield{author}{\bibinfo{person}{G~E Hinton}, \bibinfo{person}{J~L
  McClelland}, {and} \bibinfo{person}{D~E Rumelhart}.}
  \bibinfo{year}{1984}\natexlab{}.
\newblock \showarticletitle{Distributed {{Representations}}}.
\newblock In \bibinfo{booktitle}{\emph{Parallel {{Distributed Processing}}}}.
  \bibinfo{publisher}{{Carnegie-Mellon University}},
  \bibinfo{address}{Pittsburgh, PA}, \bibinfo{pages}{33}.
\newblock


\bibitem[\protect\citeauthoryear{Hitchcock}{Hitchcock}{1927}]%
        {hitchcock_expressiontensorpolyadic_1927}
\bibfield{author}{\bibinfo{person}{Frank~L. Hitchcock}.}
  \bibinfo{year}{1927}\natexlab{}.
\newblock \showarticletitle{The {{Expression}} of a {{Tensor}} or a
  {{Polyadic}} as a {{Sum}} of {{Products}}}.
\newblock \bibinfo{journal}{\emph{Journal of Mathematics and Physics}}
  \bibinfo{volume}{6}, \bibinfo{number}{1-4} (\bibinfo{date}{April}
  \bibinfo{year}{1927}), \bibinfo{pages}{164--189}.
\newblock
\showISSN{00971421}
\urldef\tempurl%
\url{https://doi.org/10.1002/sapm192761164}
\showDOI{\tempurl}


\bibitem[\protect\citeauthoryear{{Jeffrey Pennington}, {Richard Socher}, and
  {Christopher Manning}}{{Jeffrey Pennington} et~al\mbox{.}}{2014}]%
        {jeffreypennington_gloveglobalvectors_2014}
\bibfield{author}{\bibinfo{person}{{Jeffrey Pennington}},
  \bibinfo{person}{{Richard Socher}}, {and} \bibinfo{person}{{Christopher
  Manning}}.} \bibinfo{year}{2014}\natexlab{}.
\newblock \showarticletitle{{{GloVe}}: {{Global Vectors}} for {{Word
  Representation}}}. In \bibinfo{booktitle}{\emph{Proceedings of the 2014
  Conference on Empirical Methods in Natural Language Processing ({{EMNLP}})}}.
  \bibinfo{pages}{1532--1543}.
\newblock


\bibitem[\protect\citeauthoryear{Kantor and Solodovnikov}{Kantor and
  Solodovnikov}{1989}]%
        {kantor_hypercomplexnumberselementary_1989}
\bibfield{author}{\bibinfo{person}{Isai~Lvovich Kantor} {and}
  \bibinfo{person}{Aleksandr~Samuilovich Solodovnikov}.}
  \bibinfo{year}{1989}\natexlab{}.
\newblock \bibinfo{booktitle}{\emph{Hypercomplex Numbers: An Elementary
  Introduction to Algebras}}.
\newblock \bibinfo{publisher}{{Springer}}.
\newblock


\bibitem[\protect\citeauthoryear{Kingma and Ba}{Kingma and Ba}{2014}]%
        {kingma_adammethodstochastic_2014}
\bibfield{author}{\bibinfo{person}{Diederik~P. Kingma} {and}
  \bibinfo{person}{Jimmy Ba}.} \bibinfo{year}{2014}\natexlab{}.
\newblock \showarticletitle{Adam: {{A Method}} for {{Stochastic
  Optimization}}}. In \bibinfo{booktitle}{\emph{Proceedings of the 3rd
  {{International Conference}} on {{Learning Representations}} ({{ICLR}})}}.
\newblock


\bibitem[\protect\citeauthoryear{Lacroix, Usunier, and Obozinski}{Lacroix
  et~al\mbox{.}}{2018}]%
        {lacroix_canonicaltensordecomposition_2018}
\bibfield{author}{\bibinfo{person}{Timoth\'ee Lacroix},
  \bibinfo{person}{Nicolas Usunier}, {and} \bibinfo{person}{Guillaume
  Obozinski}.} \bibinfo{year}{2018}\natexlab{}.
\newblock \showarticletitle{Canonical {{Tensor Decomposition}} for {{Knowledge
  Base Completion}}}. In \bibinfo{booktitle}{\emph{Proceedings of the 35th
  {{International Conference}} on {{Machine Learning}} ({{ICML}}'18)}}.
\newblock


\bibitem[\protect\citeauthoryear{Lin, Liu, Luan, Sun, Rao, and Liu}{Lin
  et~al\mbox{.}}{2015a}]%
        {lin_modelingrelationpaths_2015}
\bibfield{author}{\bibinfo{person}{Yankai Lin}, \bibinfo{person}{Zhiyuan Liu},
  \bibinfo{person}{Huanbo Luan}, \bibinfo{person}{Maosong Sun},
  \bibinfo{person}{Siwei Rao}, {and} \bibinfo{person}{Song Liu}.}
  \bibinfo{year}{2015}\natexlab{a}.
\newblock \showarticletitle{Modeling {{Relation Paths}} for {{Representation
  Learning}} of {{Knowledge Bases}}}. In \bibinfo{booktitle}{\emph{Proceedings
  of the 2015 {{Conference}} on {{Empirical Methods}} in {{Natural Language
  Processing}}}}.
\newblock


\bibitem[\protect\citeauthoryear{Lin, Liu, Sun, Liu, and Zhu}{Lin
  et~al\mbox{.}}{2015b}]%
        {lin_learningentityrelation_2015}
\bibfield{author}{\bibinfo{person}{Yankai Lin}, \bibinfo{person}{Zhiyuan Liu},
  \bibinfo{person}{Maosong Sun}, \bibinfo{person}{Yang Liu}, {and}
  \bibinfo{person}{Xuan Zhu}.} \bibinfo{year}{2015}\natexlab{b}.
\newblock \showarticletitle{Learning {{Entity}} and {{Relation Embeddings}} for
  {{Knowledge Graph Completion}}}. In \bibinfo{booktitle}{\emph{Proceedings of
  the {{Twenty}}-{{Ninth AAAI Conference}} on {{Artificial Intelligence}}}}.
  \bibinfo{pages}{2181--2187}.
\newblock


\bibitem[\protect\citeauthoryear{Mikolov, Chen, Corrado, and Dean}{Mikolov
  et~al\mbox{.}}{2013a}]%
        {mikolov_efficientestimationword_2013}
\bibfield{author}{\bibinfo{person}{Tomas Mikolov}, \bibinfo{person}{Kai Chen},
  \bibinfo{person}{Greg Corrado}, {and} \bibinfo{person}{Jeffrey Dean}.}
  \bibinfo{year}{2013}\natexlab{a}.
\newblock \showarticletitle{Efficient {{Estimation}} of {{Word
  Representations}} in {{Vector Space}}}. In
  \bibinfo{booktitle}{\emph{{{ICLR}}'13 {{Workshop}}}}.
\newblock


\bibitem[\protect\citeauthoryear{Mikolov, Sutskever, Chen, Corrado, and
  Dean}{Mikolov et~al\mbox{.}}{2013b}]%
        {mikolov_distributedrepresentationswords_2013}
\bibfield{author}{\bibinfo{person}{Tomas Mikolov}, \bibinfo{person}{Ilya
  Sutskever}, \bibinfo{person}{Kai Chen}, \bibinfo{person}{Greg~S. Corrado},
  {and} \bibinfo{person}{Jeff Dean}.} \bibinfo{year}{2013}\natexlab{b}.
\newblock \showarticletitle{Distributed Representations of Words and Phrases
  and Their Compositionality}. In \bibinfo{booktitle}{\emph{Advances in Neural
  Information Processing Systems}}. \bibinfo{pages}{3111--3119}.
\newblock


\bibitem[\protect\citeauthoryear{{Miller, George A.}}{{Miller, George
  A.}}{1995}]%
        {millergeorgea._wordnetlexicaldatabase_1995}
\bibfield{author}{\bibinfo{person}{{Miller, George A.}}}
  \bibinfo{year}{1995}\natexlab{}.
\newblock \showarticletitle{{{WordNet}}: A Lexical Database for {{English}}}.
\newblock \bibinfo{journal}{\emph{Commun. ACM}} (\bibinfo{year}{1995}),
  \bibinfo{pages}{39--41}.
\newblock


\bibitem[\protect\citeauthoryear{Minemoto, Isokawa, Nishimura, and
  Matsui}{Minemoto et~al\mbox{.}}{2017}]%
        {minemoto_feedforwardneural_2017}
\bibfield{author}{\bibinfo{person}{Toshifumi Minemoto},
  \bibinfo{person}{Teijiro Isokawa}, \bibinfo{person}{Haruhiko Nishimura},
  {and} \bibinfo{person}{Nobuyuki Matsui}.} \bibinfo{year}{2017}\natexlab{}.
\newblock \showarticletitle{Feed Forward Neural Network with Random
  Quaternionic Neurons}.
\newblock \bibinfo{journal}{\emph{Signal Processing}} \bibinfo{volume}{C},
  \bibinfo{number}{136} (\bibinfo{year}{2017}), \bibinfo{pages}{59--68}.
\newblock
\showISSN{0165-1684}
\urldef\tempurl%
\url{https://doi.org/10.1016/j.sigpro.2016.11.008}
\showDOI{\tempurl}


\bibitem[\protect\citeauthoryear{Nickel, Tresp, and Kriegel}{Nickel
  et~al\mbox{.}}{2011}]%
        {nickel_threewaymodelcollective_2011}
\bibfield{author}{\bibinfo{person}{Maximilian Nickel}, \bibinfo{person}{Volker
  Tresp}, {and} \bibinfo{person}{Hans-Peter Kriegel}.}
  \bibinfo{year}{2011}\natexlab{}.
\newblock \showarticletitle{A {{Three}}-{{Way Model}} for {{Collective
  Learning}} on {{Multi}}-{{Relational Data}}}. In
  \bibinfo{booktitle}{\emph{Proceedings of the 28th {{International
  Conference}} on {{Machine Learning}}}}. \bibinfo{pages}{809--816}.
\newblock


\bibitem[\protect\citeauthoryear{Parcollet, Ravanelli, Morchid, Linar\`es,
  Trabelsi, Mori, and Bengio}{Parcollet et~al\mbox{.}}{2019}]%
        {parcollet_quaternionrecurrentneural_2019}
\bibfield{author}{\bibinfo{person}{Titouan Parcollet}, \bibinfo{person}{Mirco
  Ravanelli}, \bibinfo{person}{Mohamed Morchid}, \bibinfo{person}{Georges
  Linar\`es}, \bibinfo{person}{Chiheb Trabelsi}, \bibinfo{person}{Renato~De
  Mori}, {and} \bibinfo{person}{Yoshua Bengio}.}
  \bibinfo{year}{2019}\natexlab{}.
\newblock \showarticletitle{Quaternion {{Recurrent Neural Networks}}}. In
  \bibinfo{booktitle}{\emph{Proceedings of the {{International Conference}} on
  {{Learning Representations}} ({{ICLR}}'19)}}.
\newblock


\bibitem[\protect\citeauthoryear{Socher, Chen, Manning, and Ng}{Socher
  et~al\mbox{.}}{2013}]%
        {socher_reasoningneuraltensor_2013}
\bibfield{author}{\bibinfo{person}{Richard Socher}, \bibinfo{person}{Danqi
  Chen}, \bibinfo{person}{Christopher~D Manning}, {and}
  \bibinfo{person}{Andrew~Y Ng}.} \bibinfo{year}{2013}\natexlab{}.
\newblock \showarticletitle{Reasoning {{With Neural Tensor Networks}} for
  {{Knowledge Base Completion}}}. In \bibinfo{booktitle}{\emph{Advances in
  Neural Information Processing Systems}}. \bibinfo{pages}{926--934}.
\newblock


\bibitem[\protect\citeauthoryear{Trouillon, Dance, Gaussier, Welbl, Riedel, and
  Bouchard}{Trouillon et~al\mbox{.}}{2017}]%
        {trouillon_knowledgegraphcompletion_2017}
\bibfield{author}{\bibinfo{person}{Th\'eo Trouillon},
  \bibinfo{person}{Christopher~R. Dance}, \bibinfo{person}{\'Eric Gaussier},
  \bibinfo{person}{Johannes Welbl}, \bibinfo{person}{Sebastian Riedel}, {and}
  \bibinfo{person}{Guillaume Bouchard}.} \bibinfo{year}{2017}\natexlab{}.
\newblock \showarticletitle{Knowledge Graph Completion via Complex Tensor
  Factorization}.
\newblock \bibinfo{journal}{\emph{The Journal of Machine Learning Research}}
  \bibinfo{volume}{18}, \bibinfo{number}{1} (\bibinfo{year}{2017}),
  \bibinfo{pages}{4735--4772}.
\newblock


\bibitem[\protect\citeauthoryear{Trouillon, Welbl, Riedel, {Eric Gaussier}, and
  {Guillaume Bouchard}}{Trouillon et~al\mbox{.}}{2016}]%
        {trouillon_complexembeddingssimple_2016}
\bibfield{author}{\bibinfo{person}{Theo Trouillon}, \bibinfo{person}{Johannes
  Welbl}, \bibinfo{person}{Sebastian Riedel}, \bibinfo{person}{{Eric
  Gaussier}}, {and} \bibinfo{person}{{Guillaume Bouchard}}.}
  \bibinfo{year}{2016}\natexlab{}.
\newblock \showarticletitle{Complex {{Embeddings}} for {{Simple Link
  Prediction}}}. In \bibinfo{booktitle}{\emph{International {{Conference}} on
  {{Machine Learning}}}} \emph{(\bibinfo{series}{{{ICML}}'16})}.
  \bibinfo{pages}{2071--2080}.
\newblock


\bibitem[\protect\citeauthoryear{Vrande{\v c}i\'c and Kr\"otzsch}{Vrande{\v
  c}i\'c and Kr\"otzsch}{2014}]%
        {vrandecic_wikidatafreecollaborative_2014}
\bibfield{author}{\bibinfo{person}{Denny Vrande{\v c}i\'c} {and}
  \bibinfo{person}{Markus Kr\"otzsch}.} \bibinfo{year}{2014}\natexlab{}.
\newblock \showarticletitle{Wikidata: A Free Collaborative Knowledgebase}.
\newblock \bibinfo{journal}{\emph{Commun. ACM}} \bibinfo{volume}{57},
  \bibinfo{number}{10} (\bibinfo{date}{Sept.} \bibinfo{year}{2014}),
  \bibinfo{pages}{78--85}.
\newblock
\showISSN{00010782}
\urldef\tempurl%
\url{https://doi.org/10.1145/2629489}
\showDOI{\tempurl}


\bibitem[\protect\citeauthoryear{Wang, Mao, Wang, and Guo}{Wang
  et~al\mbox{.}}{2017}]%
        {wang_knowledgegraphembedding_2017}
\bibfield{author}{\bibinfo{person}{Q. Wang}, \bibinfo{person}{Z. Mao},
  \bibinfo{person}{B. Wang}, {and} \bibinfo{person}{L. Guo}.}
  \bibinfo{year}{2017}\natexlab{}.
\newblock \showarticletitle{Knowledge {{Graph Embedding}}: {{A Survey}} of
  {{Approaches}} and {{Applications}}}.
\newblock \bibinfo{journal}{\emph{IEEE Transactions on Knowledge and Data
  Engineering}} \bibinfo{volume}{29}, \bibinfo{number}{12}
  (\bibinfo{date}{Dec.} \bibinfo{year}{2017}), \bibinfo{pages}{2724--2743}.
\newblock
\showISSN{1041-4347}
\urldef\tempurl%
\url{https://doi.org/10.1109/TKDE.2017.2754499}
\showDOI{\tempurl}


\bibitem[\protect\citeauthoryear{Wang, Gemulla, and Li}{Wang
  et~al\mbox{.}}{2018}]%
        {wang_multirelationallinkprediction_2018}
\bibfield{author}{\bibinfo{person}{Yanjie Wang}, \bibinfo{person}{Rainer
  Gemulla}, {and} \bibinfo{person}{Hui Li}.} \bibinfo{year}{2018}\natexlab{}.
\newblock \showarticletitle{On {{Multi}}-{{Relational Link Prediction}} with
  {{Bilinear Models}}}. In \bibinfo{booktitle}{\emph{Thirty-{{Second AAAI
  Conference}} on {{Artificial Intelligence}}}}.
\newblock


\bibitem[\protect\citeauthoryear{Wang, Zhang, Feng, and Chen}{Wang
  et~al\mbox{.}}{2014a}]%
        {wang_knowledgegraphtext_2014}
\bibfield{author}{\bibinfo{person}{Zhen Wang}, \bibinfo{person}{Jianwen Zhang},
  \bibinfo{person}{Jianlin Feng}, {and} \bibinfo{person}{Zheng Chen}.}
  \bibinfo{year}{2014}\natexlab{a}.
\newblock \showarticletitle{Knowledge {{Graph}} and {{Text Jointly
  Embedding}}}. In \bibinfo{booktitle}{\emph{Proceedings of the 2014 Conference
  on Empirical Methods in Natural Language Processing}}.
  \bibinfo{pages}{1591--1601}.
\newblock


\bibitem[\protect\citeauthoryear{Wang, Zhang, Feng, and Chen}{Wang
  et~al\mbox{.}}{2014b}]%
        {wang_knowledgegraphembedding_2014}
\bibfield{author}{\bibinfo{person}{Zhen Wang}, \bibinfo{person}{Jianwen Zhang},
  \bibinfo{person}{Jianlin Feng}, {and} \bibinfo{person}{Zheng Chen}.}
  \bibinfo{year}{2014}\natexlab{b}.
\newblock \showarticletitle{Knowledge {{Graph Embedding}} by {{Translating}} on
  {{Hyperplanes}}}. In \bibinfo{booktitle}{\emph{{{AAAI Conference}} on
  {{Artificial Intelligence}}}}. \bibinfo{publisher}{{Citeseer}},
  \bibinfo{pages}{1112--1119}.
\newblock


\bibitem[\protect\citeauthoryear{Xiao, Huang, Hao, and Zhu}{Xiao
  et~al\mbox{.}}{2015}]%
        {xiao_transaadaptiveapproach_2015}
\bibfield{author}{\bibinfo{person}{Han Xiao}, \bibinfo{person}{Minlie Huang},
  \bibinfo{person}{Yu Hao}, {and} \bibinfo{person}{Xiaoyan Zhu}.}
  \bibinfo{year}{2015}\natexlab{}.
\newblock \showarticletitle{{{TransA}}: {{An Adaptive Approach}} for
  {{Knowledge Graph Embedding}}}. In \bibinfo{booktitle}{\emph{{{AAAI
  Conference}} on {{Artificial Intelligence}}}}.
\newblock
\showeprint[arxiv]{1509.05490}


\bibitem[\protect\citeauthoryear{Yang, Yih, He, Gao, and Deng}{Yang
  et~al\mbox{.}}{2015}]%
        {yang_embeddingentitiesrelations_2015}
\bibfield{author}{\bibinfo{person}{Bishan Yang}, \bibinfo{person}{Wen-tau Yih},
  \bibinfo{person}{Xiaodong He}, \bibinfo{person}{Jianfeng Gao}, {and}
  \bibinfo{person}{Li Deng}.} \bibinfo{year}{2015}\natexlab{}.
\newblock \showarticletitle{Embedding {{Entities}} and {{Relations}} for
  {{Learning}} and {{Inference}} in {{Knowledge Bases}}}. In
  \bibinfo{booktitle}{\emph{International {{Conference}} on {{Learning
  Representations}}}}.
\newblock


\bibitem[\protect\citeauthoryear{Zhang, Yuan, Lian, Xie, and Ma}{Zhang
  et~al\mbox{.}}{2016}]%
        {zhang_collaborativeknowledgebase_2016}
\bibfield{author}{\bibinfo{person}{Fuzheng Zhang},
  \bibinfo{person}{Nicholas~Jing Yuan}, \bibinfo{person}{Defu Lian},
  \bibinfo{person}{Xing Xie}, {and} \bibinfo{person}{Wei-Ying Ma}.}
  \bibinfo{year}{2016}\natexlab{}.
\newblock \showarticletitle{Collaborative {{Knowledge Base Embedding}} for
  {{Recommender Systems}}}. \bibinfo{publisher}{{ACM Press}},
  \bibinfo{pages}{353--362}.
\newblock
\showISBNx{978-1-4503-4232-2}
\urldef\tempurl%
\url{https://doi.org/10.1145/2939672.2939673}
\showDOI{\tempurl}


\end{thebibliography}

\newpage

\appendix
\section{On the Equivalence of the Loss Functions}
For clarity, we simplify the notation as follows. The cross-entropy loss for binary classification is written as:
\begin{align}
	\mathcal{L}_1 = - \sum{p \log \sigma(z) + (1-p) \log (1-\sigma(z))}, 
\end{align}
where $z$ is the score and $p \in \{1, 0\}$ the ground truth probability.

The softplus loss is written as:
\begin{align}
	\mathcal{L}_2 = \sum{\log(1 + e^{-y z})}, 
\end{align}
where the class label $y$ is defined by $y = 2p - 1 \in \{1, -1\}$.

Here we show that the softplus loss is equivalent to the cross-entropy loss for binary classification.

\begin{proof}
Recall that the sigmoid function is defined as:
\begin{align}
	\sigma(z) = \frac{e^z}{e^z + 1} = \frac{1}{1 + e^{-z}}. \label{eq:sigmoid}
\end{align}

This function has the interesting property that:
\begin{align}
	\sigma(-z) = \frac{1}{1 + e^z} = \frac{1 + e^z - e^z}{1 + e^z} = 1 - \frac{e^z}{1 + e^z} = 1 - \sigma(z). \label{eq:sigmoid_property_complement}
\end{align}

We also note the property of the log function:
\begin{align}
	-\log x = \log 1 - \log x = \log \frac{1}{x}. \label{eq:log_property_negative}
\end{align}

Note that in binary classification, $p$ only takes two values in $\{1, 0\}$. Taking this into account, we can rewrite $\mathcal{L}_1$ as:
\begin{align}
	\mathcal{L}_1 = &- \sum_{p=1, y=1}{p \log \sigma(z)} - \sum_{p=0, y=-1}{(1-p) \log (1-\sigma(z))} \\
	&\ \ \ \ \textit{(split positive and negative data)} \nonumber \\
	= &- \sum_{p=1, y=1}{\log \sigma(z)} - \sum_{p=0, y=-1}{\log (1-\sigma(z))} \\
	&\ \ \ \ \textit{(evaluate $p$)} \nonumber \\
	= &- \sum_{p=1, y=1}{\log \sigma(z)} - \sum_{p=0, y=-1}{\log \sigma(-z)} \\
	&\ \ \ \ \textit{(property of sigmoid function (Eq. \ref{eq:sigmoid_property_complement}))} \nonumber \\
	= &\sum_{p=1, y=1}{\log \frac{1}{\sigma(z)}} + \sum_{p=0, y=-1}{\log \frac{1}{\sigma(-z)}} \\
	&\ \ \ \ \textit{(property of log function (Eq. \ref{eq:log_property_negative}))} \nonumber \\
	= &\sum_{p=1, y=1}{\log \frac{e^z + 1}{e^z}} + \sum_{p=0, y=-1}{\log \frac{e^{-z} + 1}{e^{-z}}} \\
	= &\sum_{p=1, y=1}{\log (1 + e^{-z})} + \sum_{p=0, y=-1}{\log (1 + e^{z})} \\
	= &\sum_{p=1, y=1}{\log (1 + e^{-yz})} + \sum_{p=0, y=-1}{\log (1 + e^{-yz})} \\
	&\ \ \ \ \textit{(evaluate $y$)} \nonumber \\
	= &\sum{\log (1 + e^{-yz})} = \mathcal{L}_2 \\
	&\ \ \ \ \textit{(merge positive and negative data)} \nonumber
\end{align}
\end{proof}

From a statistical learning point of view, the cross-entropy makes a natural choice for the loss function. This loss is equivalent to the negative log-likelihood of the model on the ground truth data, following the principle of maximum likelihood estimation (MLE). For binary classification, it takes the form of $\mathcal{L}_1$ because the predicted probability is modeled by the sigmoid function. Furthermore, when the ground truth probability only takes the value 1 or 0, the softplus loss $\mathcal{L}_2$ is just a more concise form of the cross-entropy loss $\mathcal{L}_1$.

\end{document}